\documentclass{article}
\usepackage{spconf}

\usepackage{amsfonts, gensymb, mathtools, microtype}

\usepackage[main=english, ngerman]{babel}
\usepackage[T1]{fontenc}
\usepackage[utf8]{inputenc}

\usepackage[hyphens]{url}
\usepackage[colorlinks, citecolor=LimeGreen]{hyperref}
\usepackage[dvipsnames, table]{xcolor}

\usepackage{tikz}
\usetikzlibrary{
    arrows,
    backgrounds,
    calc,
    decorations,
    calligraphy,
    fit,
    positioning,
    shapes,
    spy
}

\usepackage{pgfplots}
\pgfplotsset{compat=newest}
\usepgfplotslibrary{colorbrewer}

\usepackage[hang]{caption}

\usepackage{array, diagbox, makecell, multirow, subfig, xfrac}

\usepackage[export]{adjustbox}

\usepackage{graphicx, graphbox}
\usepackage[figuresright]{rotating}
\graphicspath{{figs/}}

\usepackage{listings}
\usepackage[defaultlines=3, all]{nowidow}

\newcolumntype{P}[1]{>{ \centering  \arraybackslash }p{#1}}
\newcolumntype{Q}[1]{>{ \raggedleft \arraybackslash }p{#1}}
\newcolumntype{M}[1]{>{ \centering  \arraybackslash }m{#1}}
\newcolumntype{N}[1]{>{ \raggedleft \arraybackslash }m{#1}}
\newcolumntype{B}[1]{>{ \centering  \arraybackslash }b{#1}}
\newcolumntype{C}[1]{>{ \raggedleft \arraybackslash }b{#1}}

\usepackage{ccicons}

\usepackage{packages/hex}

\usepackage{fontawesome5}

\newcommand{\ORCID}[1]{\thinspace\textsuperscript{\href{https://orcid.org/#1}{\textcolor[HTML]{A6CE39}{\faOrcid}}}}

\newcommand{\ORCIDSchlosser}{0000-0002-0682-4284} % ORCID Tobias Schlosser
\newcommand{\ORCIDBeuth}{0000-0001-5482-9787}     % ORCID Frederik Beuth
\newcommand{\ORCIDKowerko}{0000-0002-4538-7814}   % ORCID Danny Kowerko

\usepackage{fancyhdr, lastpage}
\fancypagestyle{plain}{\fancyfoot[C]{\thepage/\pageref{LastPage}}}
\begin{document}

%%%%%%%%%%%%%%%%%%%%%%%%%%%%%%%%%%%% Title %%%%%%%%%%%%%%%%%%%%%%%%%%%%%%%%%%%%%
\title{Biologically Inspired Hexagonal Deep Learning for \\ Hexagonal Image Generation}

\name{
    Tobias Schlosser\ORCID{\ORCIDSchlosser},
    Frederik Beuth\ORCID{\ORCIDBeuth}, and
    Danny Kowerko\ORCID{\ORCIDKowerko}
}

\address{
    Junior Professorship of Media Computing, \\
    Chemnitz University of Technology, \\
    09107 Chemnitz, Germany, \\
    \texttt{\small \{firstname.lastname\}@cs.tu-chemnitz.de}
}
%%%%%%%%%%%%%%%%%%%%%%%%%%%%%%%%%%%% Title %%%%%%%%%%%%%%%%%%%%%%%%%%%%%%%%%%%%%

\maketitle

\begin{abstract}
    %%%%%%%%%%%%%%%%%%%%%%%%%%%%%%%%%%%%% Text %%%%%%%%%%%%%%%%%%%%%%%%%%%%%%%%%%%%%
    Whereas conventional state-of-the-art image processing systems of recording and output devices almost exclusively utilize square arranged methods, biological models, however, suggest an alternative, evolutionarily-based structure. Inspired by the human visual perception system, hexagonal image processing in the context of machine learning offers a number of key advantages that can benefit both researchers and users alike. The hexagonal deep learning framework Hexnet leveraged in this contribution serves therefore the generation of hexagonal images by utilizing hexagonal deep neural networks (H-DNN). As the results of our created test environment show, the proposed models can surpass current approaches of conventional image generation. While resulting in a reduction of the models' complexity in the form of trainable parameters, they furthermore allow an increase of test rates in comparison to their square counterparts.
    %%%%%%%%%%%%%%%%%%%%%%%%%%%%%%%%%%%%% Text %%%%%%%%%%%%%%%%%%%%%%%%%%%%%%%%%%%%%
\end{abstract}


\section{Introduction and motivation}

%%%%%%%%%%%%%%%%%%%%%%%%%%%%%%%%%%%%% Text %%%%%%%%%%%%%%%%%%%%%%%%%%%%%%%%%%%%%
While the structure and functionality of artificial neural networks is inspired by biological processes, they are also limited by their underlying structure. Due to the current state of the art of recording and output devices, mostly square structures are used, which also significantly restrict subsequent image processing systems \cite{Staunton1990}.

With the advancements of recent years, machine learning and deep neural networks (DNN) are becoming increasingly important. In order to meet the ever-increasing demands of more complex problems and systems, novel application areas and larger data sets are being investigated and developed \cite{Krizhevsky2012}. Entailing their steadily increasing diversity \cite{Szegedy2015}, novel architectures for artificial neural networks emerge that exceed and even supersede conventional Cartesian-based approaches, including spherical and non-Euclidean manifolds \cite{Bronstein2017}.

In comparison to the current state of the art, the human visual perception system suggests an alternative, evolutionarily-based structure. To allow an efficient processing of incoming signals \cite{Middleton2005}, the retina of the human eye displays a particularly strong hexagonal arrangement of sensory cells \cite{Curcio1987}. The image in the visual cortex is following the reduction of the perceived information to the nerve fibers projected retinotopic, whereas neighboring structures are preserved and processed through the following areas of the brain \cite{Hubel1968}.

\begin{figure}[tb]
    \centering

    \subfloat[Square lattice format]{
        \scalebox{0.67}{\begin{tikzpicture}[r/.style={draw, rectangle, minimum size=12.5mm}]
            \node[r, fill=Red!95] (r1) at (-1.25,  1.25) {};
            \node[r, fill=Red!85] (r2) at ( 0,     1.25) {};
            \node[r]              (r3) at ( 1.25,  1.25) {};
            \node[r]              (r4) at (-1.25,  0)    {};
            \node[r]              (r5) at ( 0,     0)    {};
            \node[r]              (r6) at ( 1.25,  0)    {};
            \node[r]              (r7) at (-1.25, -1.25) {};
            \node[r]              (r8) at ( 0,    -1.25) {};
            \node[r]              (r9) at ( 1.25, -1.25) {};

            \node[draw, circle, minimum size=2mm, inner sep=0, semithick] (c) at (0, 0) {};

            \draw[->, semithick] (c) -- node[left, yshift=-3mm] {$\sqrt{2}$} (-1.25, 1.25);
            \draw[->, semithick] (c) -- node[right]             {1}          ( 0,    1.25);

            \draw[->, shorten <=-1mm] (-3.66, 0.5) -- (-2.66, 0.5) node[right] {$x$};
            \draw[->, shorten <=-1mm] (-3.66, 0.5) -- (-3.66, 1.5) node[above] {$y$};
        \end{tikzpicture}}}
    \quad
    \subfloat[Hexagonal lattice format]{
        \scalebox{0.67}{\begin{tikzpicture}[h/.style={draw, shape=regular polygon, regular polygon sides=6, rotate=30, minimum size=15mm}]
            \node[h]                 (0) at ( 0,        0)     {};
            \node[h]                 (1) at ( 1.3,      0)     {};
            \node[h, fill=LimeGreen] (2) at ( 1.3 / 2,  1.125) {};
            \node[h, fill=LimeGreen] (3) at (-1.3 / 2,  1.125) {};
            \node[h]                 (4) at (-1.3,      0)     {};
            \node[h]                 (5) at (-1.3 / 2, -1.125) {};
            \node[h]                 (6) at ( 1.3 / 2, -1.125) {};

            \node[draw, circle, minimum size=2mm, inner sep=0, semithick] (c) at (0, 0) {};

            \draw[->, semithick] (c) -- node[right] {1} ( 1.3 / 2, 1.125);
            \draw[->, semithick] (c) -- node[left]  {1} (-1.3 / 2, 1.125);

            \draw[->, shorten <=-1mm] (2.66, 0.5) -- (3.66,           0.5)         node[right] {$x$};
            \draw[->, shorten <=-1mm] (2.66, 0.5) -- (2.66 - 1.3 / 2, 0.5 + 1.125) node[above] {$y$};
            \draw[->, shorten <=-1mm] (2.66, 0.5) -- (2.66 + 1.3 / 2, 0.5 + 1.125) node[above] {$z$};
        \end{tikzpicture}}}

    \caption{Lattice format comparison. \cite{Schlosser2019_ICMLA}}
    \label{figure:lattice_formats}
\end{figure}
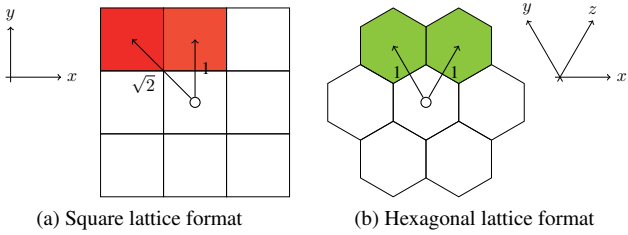

The rise and advancement of hexagonal structures and image processing systems emerges therefore as an evolutionarily motivated approach. Following this principle, hexagonal structures (Fig.~\ref{figure:lattice_formats}) feature a number of decisive advantages. These include, inter alia, the homogeneity of the hexagonal lattice format, whereby the equidistance and uniqueness of neighborhood as well as an increased radial symmetry are given. Together with a by $13.4$~\% increased transformation efficiency, hexagonal representations allow the storage of larger amounts of data on the same number of sampling points \cite{Petersen1962}. Moreover, they result in a reduction of computation times, less quantization errors, and an increased efficiency in programming \cite{Golay1969}.
%%%%%%%%%%%%%%%%%%%%%%%%%%%%%%%%%%%%% Text %%%%%%%%%%%%%%%%%%%%%%%%%%%%%%%%%%%%%

\subsection{Application areas and related work}

%%%%%%%%%%%%%%%%%%%%%%%%%%%%%%%%%%%%% Text %%%%%%%%%%%%%%%%%%%%%%%%%%%%%%%%%%%%%
Prominent application areas of hexagonal image processing systems often originate from the research fields of observation, experiment, and simulation in ecology \cite{Birch2007}, the design of geodesic grid systems \cite{Sahr2003}, sensor-based image processing and remote sensing \cite{Ambrosio2001}, medical imaging \cite{Neeser2000}, and image synthesis \cite{Theussl2001}. First experimental results for supply demand forecasting \cite{Ke2018} and the analysis of atmospheric telescope data for event detection and classification \cite{Shilon2019} explain the need for novel generative approaches within these emerging application areas.

In the context of hexagonal image processing and deep learning based systems, this entails not only principles from both domains, but also the development of operations and layers for hexagonal models \cite{Steppa2019} as well as the underlying addressing scheme and the following processing steps.
%%%%%%%%%%%%%%%%%%%%%%%%%%%%%%%%%%%%% Text %%%%%%%%%%%%%%%%%%%%%%%%%%%%%%%%%%%%%

\subsection{Contribution of hexagonal image generation}

%%%%%%%%%%%%%%%%%%%%%%%%%%%%%%%%%%%%% Text %%%%%%%%%%%%%%%%%%%%%%%%%%%%%%%%%%%%%
This contribution is meant as a general application-oriented approach to hexagonal image generation by synthesizing the advantages of the research fields of hexagonal image processing with deep learning. Following, the by \textit{Schlosser et al.} \cite{Schlosser2019_ICMLA} introduced hexagonal deep learning framework Hexnet will therefore be utilized as a first approach to hexagonal image generation using hexagonal deep neural networks.

As of the current state of art, true hexagonal images have to be captured using a hexagonal sensor. However, the corresponding hardware is either rare or limited in its applicability. Suitable alternatives include therefore the transformation or synthesis of hexagonal images \cite{Steppa2019}. While hexagonal approaches for image generation can benefit both researchers and users alike, we will provide the necessary tools to ease the application, development, and evaluation of said approaches. The related developed architectures, procedures, and test results are made publicly available and are found on the project page of Hexnet\footnote{\url{https://github.com/TSchlosser13/Hexnet}}.
%%%%%%%%%%%%%%%%%%%%%%%%%%%%%%%%%%%%% Text %%%%%%%%%%%%%%%%%%%%%%%%%%%%%%%%%%%%%

\section{Fundamentals and methods}

%%%%%%%%%%%%%%%%%%%%%%%%%%%%%%%%%%%%% Text %%%%%%%%%%%%%%%%%%%%%%%%%%%%%%%%%%%%%
As an essential element of hexagonal image processing systems, the underlying addressing scheme for transformation and visualization as well as the the necessary related processing steps have to be determined. Following, we will introduce our hexagonal addressing scheme as identified for its applicability. Based on the proposed architecture, we will then introduce the principles of hexagonal deep learning.
%%%%%%%%%%%%%%%%%%%%%%%%%%%%%%%%%%%%% Text %%%%%%%%%%%%%%%%%%%%%%%%%%%%%%%%%%%%%

\subsection{The hexagonal addressing scheme}

%%%%%%%%%%%%%%%%%%%%%%%%%%%%%%%%%%%%% Text %%%%%%%%%%%%%%%%%%%%%%%%%%%%%%%%%%%%%
Currently deployed hexagonal addressing schemes are often inspired by the so-called 1-D spiral architecture addressing scheme (SAA) \cite{Burt1980}, which established itself as the preferred addressing scheme for hexagonal image processing systems. However, one drawback of SAA originates from the reduced locality and increased complexity when transforming square images into a hexagonal representation. Pseudohexagonal addressing schemes are in contrary easier to implement \cite{Fitz1996}.

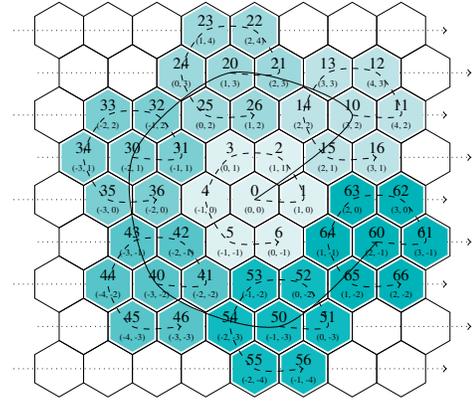
\begin{figure}[tb]
    \centering

    \scalebox{0.75}{\begin{tikzpicture}[
     h/.style={shape=regular polygon, regular polygon sides=6, shape border rotate=30, minimum size=10mm},
     align=center, font=\small, inner sep=0]
        \drawhexarray{0}{draw}{0}{1}{7}{8}
        \drawhexarraysaaofsecondorder{0}{draw=none, scale=0.66}{4}{4}{0}{0}{0}{BlueGreen}
    \end{tikzpicture}}

    \caption{Hexagonal addressing scheme construction. \cite{Schlosser2019_ICMLA}}
    \label{figure:addressing_scheme}
\end{figure}

This contribution leverages a hybrid approach that combines a 1-D linewise architecture with SAA (Fig.~\ref{figure:addressing_scheme}). Following the hexagonal addressing scheme, the from the construction process derived hierarchical hexagonal structures are exploited based on the properties of their pyramidal decompositions, also called septrees or Hexarrays. To allow an efficient implementation of the pseudohexagonal addressing scheme, the shifted Cartesian coordinates are stored for every second row \cite{Her1994}. For transformation, the Hexarray is centered over the given image and interpolated using either nearest-neighbor, bilinear, bicubic, or more complex interpolation approaches, such as hexagonal splines \cite{Her1994}.
%%%%%%%%%%%%%%%%%%%%%%%%%%%%%%%%%%%%% Text %%%%%%%%%%%%%%%%%%%%%%%%%%%%%%%%%%%%%

\subsection{Conventional and hexagonal deep neural networks}

%%%%%%%%%%%%%%%%%%%%%%%%%%%%%%%%%%%%% Text %%%%%%%%%%%%%%%%%%%%%%%%%%%%%%%%%%%%%
Whereas conventional deep neural networks are almost exclusively subdivided based on square lattice formats, hereinafter referred to as square DNNs (S-DNN), they are often divided into a set of alternating layers, which consist of convolutional, pooling, and fully connected layers. Following this principle, hexagonal deep neural networks (H-DNN) are based on the introduced hexagonal addressing scheme.
%%%%%%%%%%%%%%%%%%%%%%%%%%%%%%%%%%%%% Text %%%%%%%%%%%%%%%%%%%%%%%%%%%%%%%%%%%%%

\subsection{Assessing the quality of hexagonal representations}

%%%%%%%%%%%%%%%%%%%%%%%%%%%%%%%%%%%%% Text %%%%%%%%%%%%%%%%%%%%%%%%%%%%%%%%%%%%%
To quantify the information content and image quality of square and hexagonal lattice format based representations, the hereinafter so-called transformation efficiencies are obtained \cite{Schlosser2019_ICMLA}. Given the original square lattice format based image $S$ and its hexagonally sampled image $H$, the absolute error is derived by weighting the from projection of $S$ onto $H$ obtained subareas $a \in A$ in subpixel resolution. Following this principle, the computation of the mean squared error (MSE) is shown in~\eqref{equation:MSE}, whereas the subareas area is denoted by $|a|$.
%%%%%%%%%%%%%%%%%%%%%%%%%%%%%%%%%%%%% Text %%%%%%%%%%%%%%%%%%%%%%%%%%%%%%%%%%%%%

\begin{equation}
    \mathit{MSE} = \frac{1}{|A|} \cdot \sum\limits_{a \in A} |a| \cdot [S(a) - H(a)]^2
    \label{equation:MSE}
\end{equation}

\begin{figure*}[tb]
    \centering

    \subfloat[H-SWWAE]{
        \scalebox{0.7}{\begin{tikzpicture}[
         h/.style={shape=regular polygon, regular polygon sides=6, shape border rotate=30, minimum size=1cm},
         r/.style={draw, rectangle, minimum size=1.5cm},
         font=\normalsize]
            \node[r] (n1) at (0,  5)   {};
            \node[r] (n2) at (0,  2.5) {};
            \node[r] (n3) at (0,  0)   {};
            \node[r] (n4) at (0, -2.5) {};
            \node[r] (n5) at (0, -5)   {};

            \begin{scope}[yshift=5cm, scale=0.125, transform shape]
                \foreach \i in {0, ..., 4} {
                    \node[r, fill=white, minimum size=6cm] at (-0.66+0.33*\i, 0.66-0.33*\i) {};
                }
            \end{scope}

            \begin{scope}[yshift=2.5cm, scale=0.125, transform shape]
                \foreach \i in {0, ..., 4} {
                    \node[r, fill=white, minimum size=3cm] at (-0.66+0.33*\i, 0.66-0.33*\i) {};
                }
            \end{scope}

            \draw (0, 0) circle (0.125);

            \begin{scope}[yshift=-2.5cm, scale=0.125, transform shape]
                \foreach \i in {0, ..., 4} {
                    \ha{0}{draw, fill=white, shift={(-0.66cm+0.33*\i cm, 0.66cm-0.33*\i cm)}}{0}{0}{2}{2}
                }
            \end{scope}

            \begin{scope}[yshift=-5cm, scale=0.125, transform shape]
                \foreach \i in {0, ..., 4} {
                    \ha{0}{draw, fill=white, shift={(-0.66cm+0.33*\i cm, 0.66cm-0.33*\i cm)}}{0}{0}{5}{5}
                }
            \end{scope}

            \draw[->] (n1) -- (n2);
            \draw[->] (n2) -- (n3);
            \draw[->] (n3) -- (n4);
            \draw[->] (n4) -- (n5);

            \node[r, very thin, fit=(n1)(n5), inner sep=0.5cm] (fn15) {};

            \node[r, left=2cm of n1] (n1e) {\includegraphics[width=2cm]{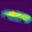}};
            \node[r, left=2cm of n2] (n2e) {\includegraphics[width=2cm]{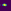}};
            \node[r, left=2cm of n3] (n3e) {$(x_1, \dots, x_n)$};
            \node[r, left=2cm of n4] (n4e) {\includegraphics[width=2cm]{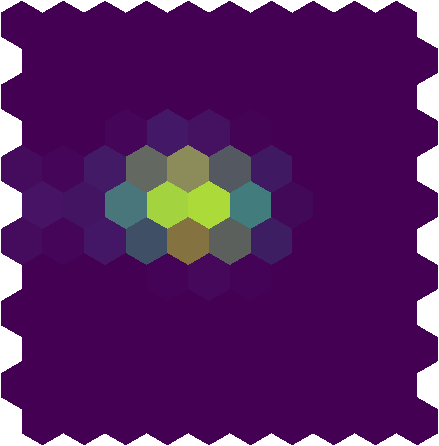}};
            \node[r, left=2cm of n5] (n5e) {\includegraphics[width=2cm]{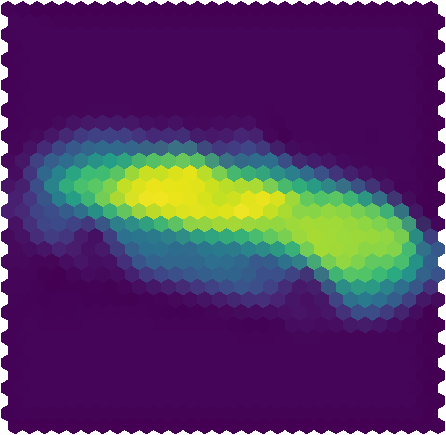}};

            \draw[dashdotted] (n1e) -- (n1);
            \draw[dashdotted] (n2e) -- (n2);
            \draw[dashdotted] (n3e) -- (n3);
            \draw[dashdotted] (n4e) -- (n4);
            \draw[dashdotted] (n5e) -- (n5);

            \draw[decorate, decoration={brace, amplitude=0.25cm}] ([xshift=1.5cm]n1.north) -- ([xshift=1.5cm]n2.south) node[midway, xshift=0.5cm, rotate=90, anchor=north] {square pooling};
            \draw[decorate, decoration={brace, amplitude=0.25cm}] ([xshift=1.5cm]n3.north) -- ([xshift=1.5cm]n3.south) node[midway, xshift=0.5cm, rotate=90, anchor=north] {core feature space};
            \draw[decorate, decoration={brace, amplitude=0.25cm}] ([xshift=1.5cm]n4.north) -- ([xshift=1.5cm]n5.south) node[midway, xshift=0.5cm, rotate=90, anchor=north] {hexagonal upsampling};
        \end{tikzpicture}}}
    \hspace{2cm}
    \subfloat[H-ACGAN]{
        \scalebox{0.7}{\begin{tikzpicture}[
         r/.style={draw, rectangle, minimum size=1cm},
         t/.style={draw, trapezium, minimum size=1cm},
         font=\normalsize]
            \node[r] (n1) at (-1,  5) {real};
            \node[r] (n2) at (-1,  3) {fake};
            \node[r] (n3) at ( 1,  3) {\itshape class};
            \node[t] (n4) at ( 0,  1) {\Large\bfseries\itshape D};
            \node[r] (n5) at (-1, -1) {$x$};
            \node[r] (n6) at ( 1, -1) {$G(z)$};
            \node[t] (n7) at ( 1, -3) {\Large\bfseries\itshape G};
            \node[r] (n8) at (-1, -5) {$c$};
            \node[r] (n9) at ( 1, -5) {$z$};

            \node[r, very thin, fit=(n1)(n2), inner sep=0.25cm] (fn_1_2) {};

            \node (n10) at ($(n2)!0.5!(n3)$)  {};
            \node (n11) at ($(n10)!0.5!(n4)$) {};
            \node (n12) at ($(n5)!0.5!(n6)$)  {};
            \node (n13) at ($(n4)!0.5!(n12)$) {};
            \node (n14) at ($(n7)!0.5!(n9)$)  {};

            \draw[->] (n4) -- (n11.center) -| (fn_1_2);
            \draw[->] (n4) -| (n3);
            \draw[->] (n5) |- (n13.center) -- (n4);
            \draw     (n6) |- (n13.center);
            \draw[->] (n7) -- (n6);
            \draw[->] ([xshift=-0.16cm]n8.north) -- ([xshift=-0.16cm]n5.south);
            \draw[->, dashed] ([xshift=0.16cm]n8.north) |- (n14.center);
            \draw[->] (n9) -- (n7);

            \node[r, very thin, fit=(fn_1_2)(n7)(n9), inner sep=0.5cm] (fn__fn_1_2__7_9) {};

            \node[r, left=2cm of n1, yshift=0.5cm]  (n1e) {\includegraphics[width=2cm]{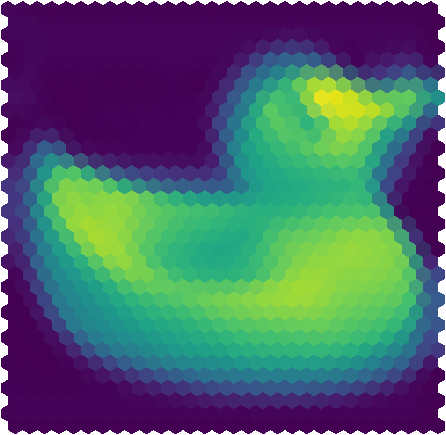}};
            \node[r, left=2cm of n2, yshift=-0.5cm] (n2e) {\includegraphics[width=2cm]{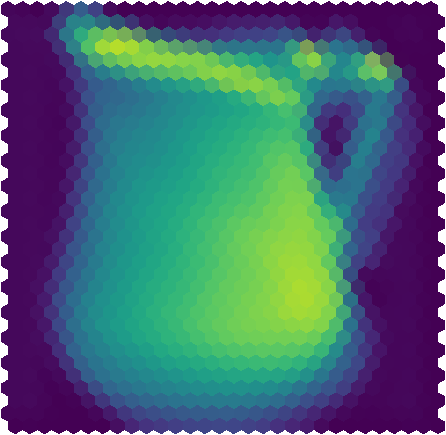}};
            \node[r, right = 2cm of n3] (n3e) {duck / jug};
            \node[r, left  = 2cm of n5] (n5e) {\includegraphics[width=2cm]{{generative_models/duck}.png}};
            \node[r, right = 2cm of n6] (n6e) {\includegraphics[width=2cm]{{generative_models/jug}.png}};
            \node[r, left  = 2cm of n8] (n8e) {duck};
            \node[r, right = 2cm of n9] (n9e) {$(z_1, \dots, z_n)$};

            \draw[dashdotted] (n1e) -- (n1);
            \draw[dashdotted] (n2e) -- (n2);
            \draw[dashdotted] (n3)  -- (n3e);
            \draw[dashdotted] (n5e) -- (n5);
            \draw[dashdotted] (n6)  -- (n6e);
            \draw[dashdotted] (n8e) -- (n8);
            \draw[dashdotted] (n9)  -- (n9e);

            \node[below=0.25cm of n3e] (n3el) {class samples};
            \node[below=0.25cm of n8e] (n8el) {class sample};
            \node[below=0.25cm of n9e] (n9el) {latent vector sample};

            \draw[decorate, decoration={brace, amplitude=0.25cm}] ([xshift=0.5cm] n6e.east |- fn_1_2.north) -- ([xshift=0.5cm] n6e.east |- n4.south)  node[midway, xshift=0.5cm, rotate=90, anchor=north, align=center] {hexagonal image \\ discrimination};
            \draw[decorate, decoration={brace, amplitude=0.25cm}] ([xshift=0.5cm] n6e.east |- n6e.north)    -- ([xshift=0.5cm] n6e.east |- n9e.south) node[midway, xshift=0.5cm, rotate=90, anchor=north, align=center] {hexagonal image \\ sampling / generation};
        \end{tikzpicture}}}

    \caption{Exemplary models for hexagonal image generation. \textbf{a)} Hexagonal Deeply Stacked Residual What-Where Autoencoder (H-SWWAE). \textbf{b)} Hexagonal Auxiliary Classifier Generative Adversarial Network (H-ACGAN).}
    \label{figure:generative_models}
\end{figure*}

\begin{table}[tb]
    \renewcommand{\arraystretch}{1.1}
    \centering

    \scalebox{0.9}{\begin{tabular}{|M{2cm}|M{2cm}|M{1.9cm}|M{1.9cm}|}
        \hline
        Model & Convolutional kernel sizes & Input shape(s) & Output shape(s) \\
        \hline
        \noalign{\vskip 2pt}

        \hline
        Encoder & $3 \times 3$ / $7^1$ & $(32, 32, 3)$ & $(1, 1, 128)$ \\
        \hline
        Decoder & $3 \times 3$ / $7^1$ & $(1, 1, 128)$ & $(32, 32, 3)$ / $(34, 30, 3)$ \\
        \hline
    \end{tabular}}

    \caption{S- / H-SWWAE layer configuration.}
    \label{table:SWWAE_model_configuration}
\end{table}

\begin{table}[tb]
    \renewcommand{\arraystretch}{1.1}
    \centering

    \scalebox{0.9}{\begin{tabular}{|M{2cm}|M{2cm}|M{1.9cm}|M{1.9cm}|}
        \hline
        Model & Convolutional kernel sizes & Input shape(s) & Output shape(s) \\
        \hline
        \noalign{\vskip 2pt}

        \hline
        Generator     & $3\times 3$ / $7^1$ & $(8192)$ & $(32, 32, 3)$ / $(34, 30, 3)$ \\
        \hline
        Discriminator & $3\times 3$ / $7^1$ & $(32, 32, 3)$ / $(34, 30, 3)$ & $(10)$ \\
        \hline
    \end{tabular}}

    \caption{S- / H-ACGAN layer configuration.}
    \label{table:ACGAN_model_configuration}
\end{table}

\section{Implementation of the hexagonal deep learning Framework}

%%%%%%%%%%%%%%%%%%%%%%%%%%%%%%%%%%%%% Text %%%%%%%%%%%%%%%%%%%%%%%%%%%%%%%%%%%%%
To allow the utilization of hexagonal deep learning models, we based our framework on the currently most commonly in research and application deployed machine learning framework TensorFlow\footnote{\url{https://www.tensorflow.org/}, \textit{version 2.1.0}} with Keras as its front end.
%%%%%%%%%%%%%%%%%%%%%%%%%%%%%%%%%%%%% Text %%%%%%%%%%%%%%%%%%%%%%%%%%%%%%%%%%%%%

\subsection{Hexagonal layers}

%%%%%%%%%%%%%%%%%%%%%%%%%%%%%%%%%%%%% Text %%%%%%%%%%%%%%%%%%%%%%%%%%%%%%%%%%%%%
Hexagonal layers for H-DNNs encompass as compared to conventional DNNs not only hexagonal equivalents for convolutional and pooling layers, but also, e.g., dense and batch normalization layers. Since hexagonal kernels are based on the introduced hexagonal addressing scheme, especially convolutional layers and their inverse are implemented intuitively by shifting SAA over the given input. The hexagonal pooling layer is then either realized based on conventional pooling operations or as inspired by the scaling of Hexarrays \cite{Middleton2005}, whereas the sub-Hexarray orders are given by the proposed addressing scheme. To solve the from two different sub-Hexarrays resulting offset mapping with or without strides corresponding to the current offsets, the mapping problem has to be approached in a more general manner. The mapping-based linear assignment problem is then solved by minimizing the resulting $\ell^2$ norm based cost \cite{Schlosser2019_ICMLA}.
%%%%%%%%%%%%%%%%%%%%%%%%%%%%%%%%%%%%% Text %%%%%%%%%%%%%%%%%%%%%%%%%%%%%%%%%%%%%

\subsection{Models for hexagonal image generation}

%%%%%%%%%%%%%%%%%%%%%%%%%%%%%%%%%%%%% Text %%%%%%%%%%%%%%%%%%%%%%%%%%%%%%%%%%%%%
The in this contribution proposed models for hexagonal image generation are divided into two groups: \textbf{a)} H-DNNs for square lattice format based images and \textbf{b)} hexagonal lattice format based ones. Former include therefore conventional square data sets, while latter include transformed, synthesized \cite{Steppa2019}, or true hexagonal images \cite{Shilon2019}.

Inspired by \textit{Zhao et al.'s} \cite{Zhao2015} and \textit{Odena et al.'s} \cite{Odena2017} approaches to Stacked What-Where Autoencoders (SWWAE) with residual learning and Auxiliary Classifier Generative Adversarial Networks (ACGAN), their exemplary hexagonal counterparts are shown in Fig.~\ref{figure:generative_models}. Every conventional layer, including not only convolutional and pooling layers, but also, i.a., upsampling layers, was substituted by its hexagonal equivalent for either \textbf{a)} only the model's decoder (H-SWWAE) or \textbf{b)} the whole model (H-ACGAN).
%%%%%%%%%%%%%%%%%%%%%%%%%%%%%%%%%%%%% Text %%%%%%%%%%%%%%%%%%%%%%%%%%%%%%%%%%%%%

\begin{figure}[tb]
    \newcommand{\width}{0.042\textwidth}
    \centering

    % MNIST square

    \includegraphics[width=\width]{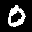}
    \includegraphics[width=\width]{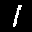}
    \includegraphics[width=\width]{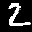}
    \includegraphics[width=\width]{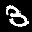}
    \includegraphics[width=\width]{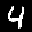}
    \hspace{0.2cm}
    \includegraphics[width=\width]{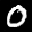}
    \includegraphics[width=\width]{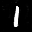}
    \includegraphics[width=\width]{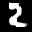}
    \includegraphics[width=\width]{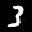}
    \includegraphics[width=\width]{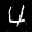}
    \vspace{0.1cm}

    \includegraphics[width=\width]{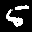}
    \includegraphics[width=\width]{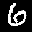}
    \includegraphics[width=\width]{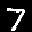}
    \includegraphics[width=\width]{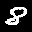}
    \includegraphics[width=\width]{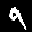}
    \hspace{0.2cm}
    \includegraphics[width=\width]{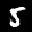}
    \includegraphics[width=\width]{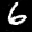}
    \includegraphics[width=\width]{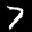}
    \includegraphics[width=\width]{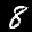}
    \includegraphics[width=\width]{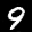}
    \vspace{0.2cm}

    % MNIST hexagonal

    \includegraphics[width=\width]{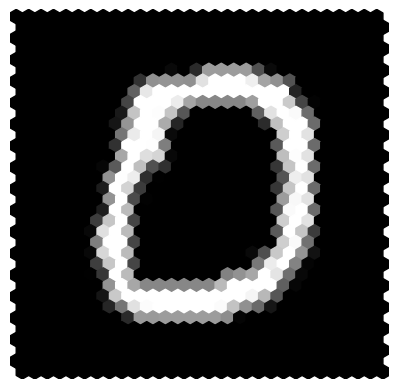}
    \includegraphics[width=\width]{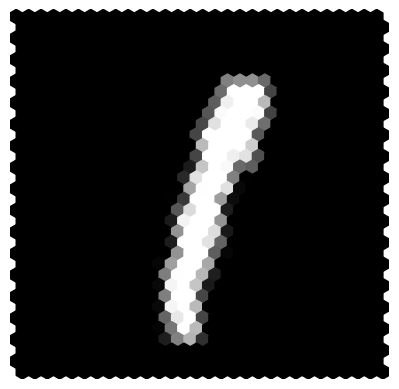}
    \includegraphics[width=\width]{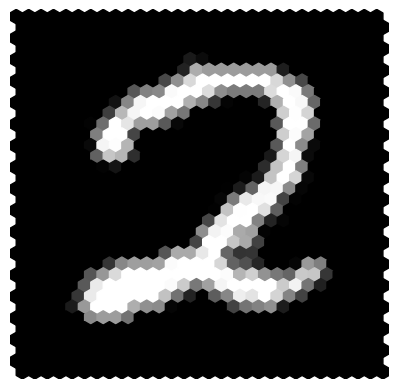}
    \includegraphics[width=\width]{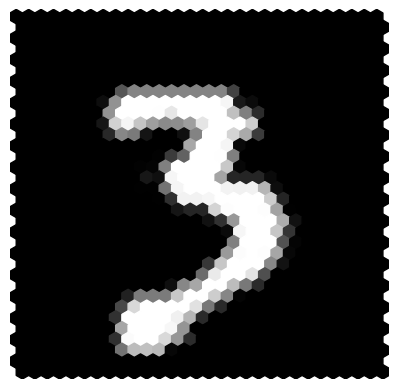}
    \includegraphics[width=\width]{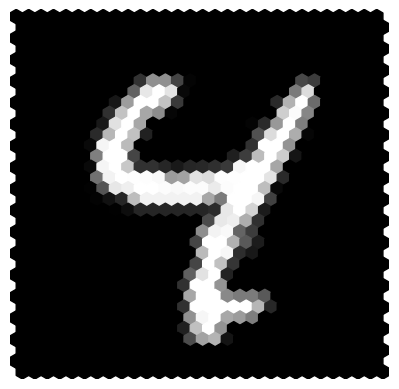}
    \hspace{0.2cm}
    \includegraphics[width=\width]{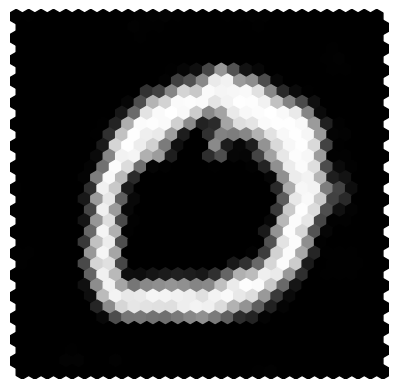}
    \includegraphics[width=\width]{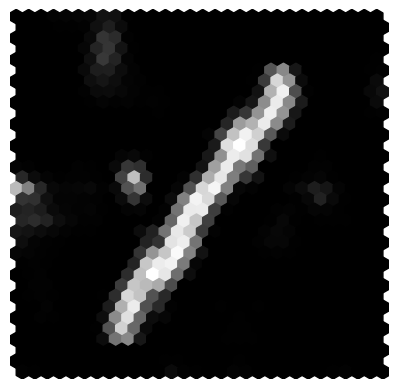}
    \includegraphics[width=\width]{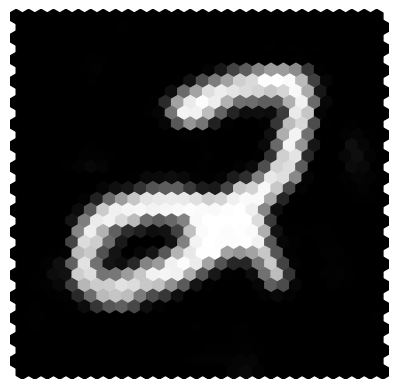}
    \includegraphics[width=\width]{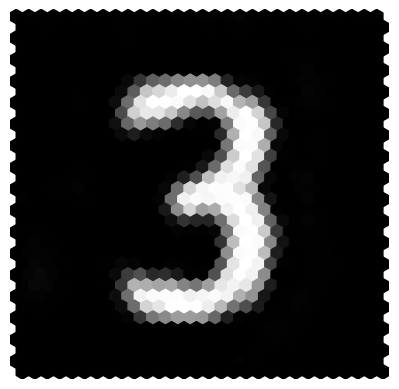}
    \includegraphics[width=\width]{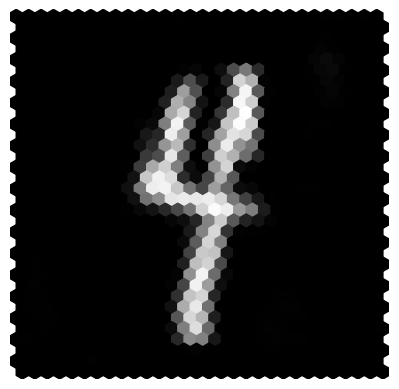}
    \vspace{0.1cm}

    \includegraphics[width=\width]{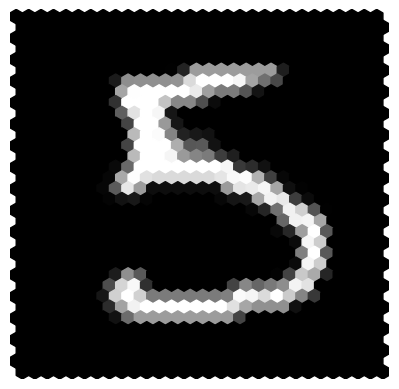}
    \includegraphics[width=\width]{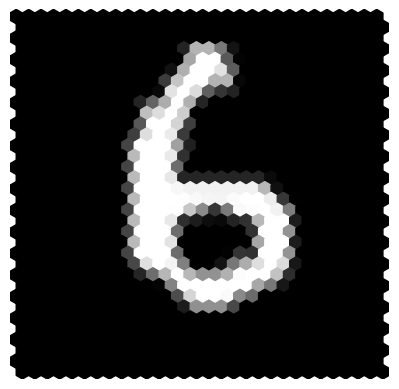}
    \includegraphics[width=\width]{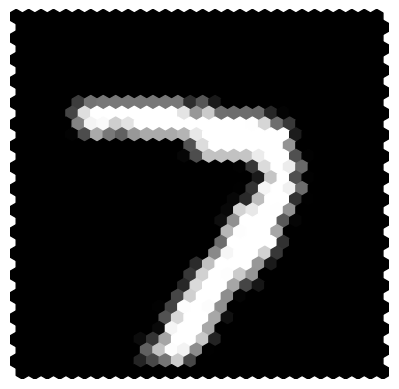}
    \includegraphics[width=\width]{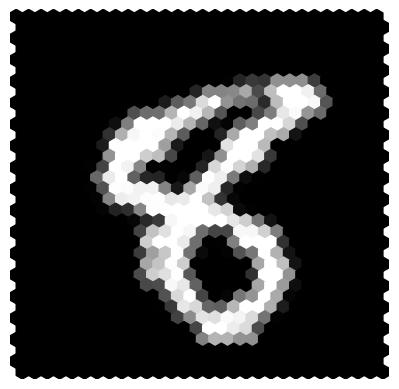}
    \includegraphics[width=\width]{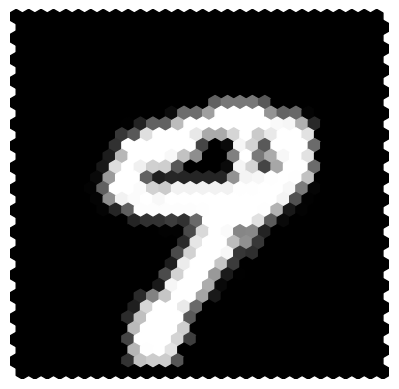}
    \hspace{0.2cm}
    \includegraphics[width=\width]{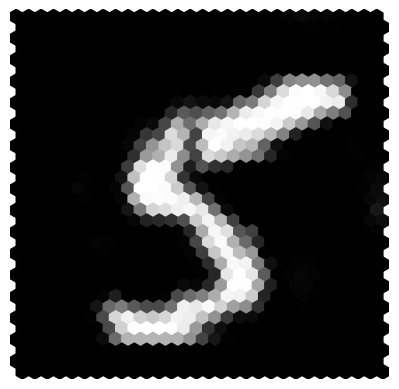}
    \includegraphics[width=\width]{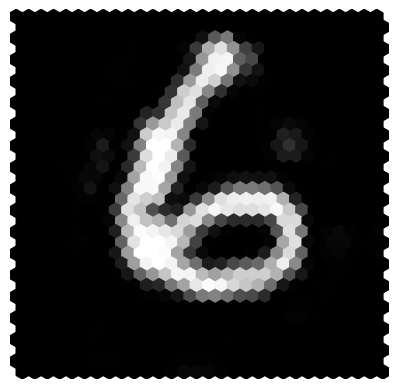}
    \includegraphics[width=\width]{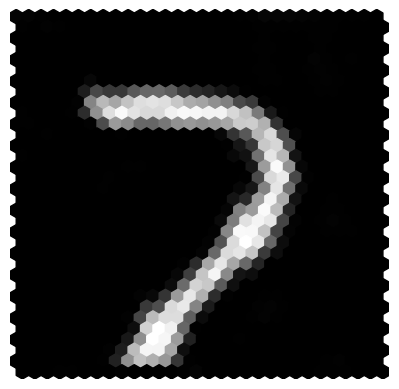}
    \includegraphics[width=\width]{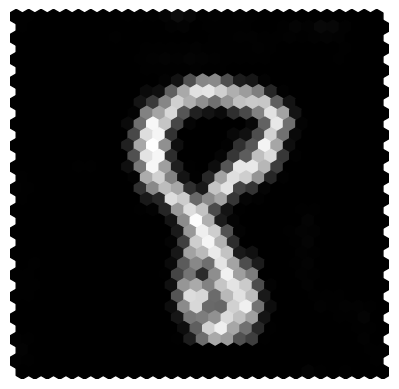}
    \includegraphics[width=\width]{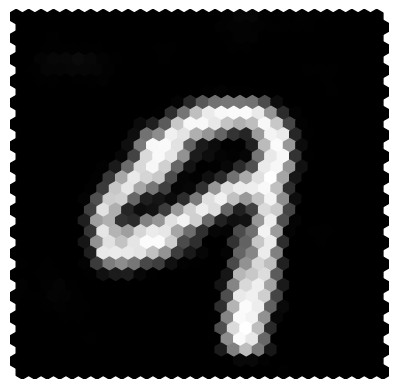}
    \vspace{-0.2cm}

    \rule[0.1cm]{\linewidth}{0.5pt}

    % CIFAR-10 hexagonal

    \includegraphics[width=\width]{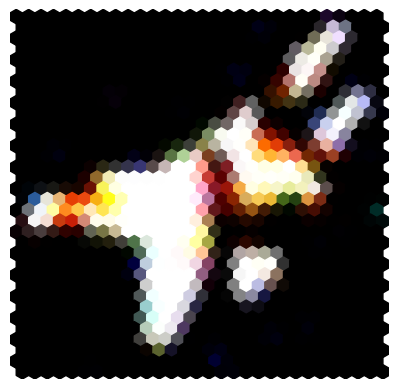}
    \includegraphics[width=\width]{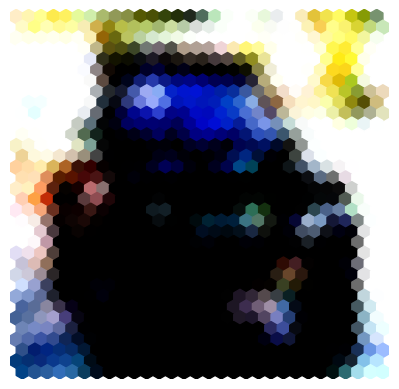}
    \includegraphics[width=\width]{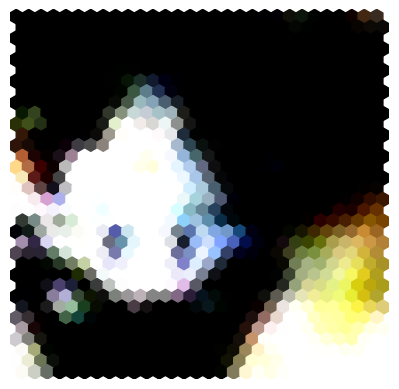}
    \includegraphics[width=\width]{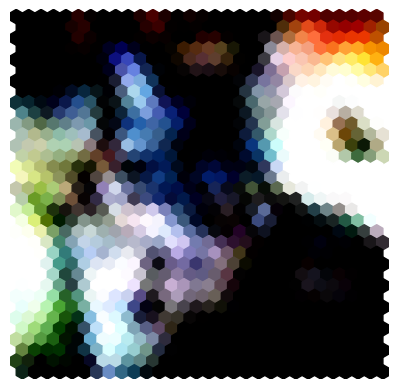}
    \includegraphics[width=\width]{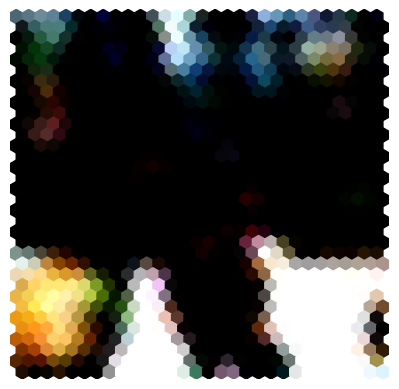}
    \hspace{0.2cm}
    \includegraphics[width=\width]{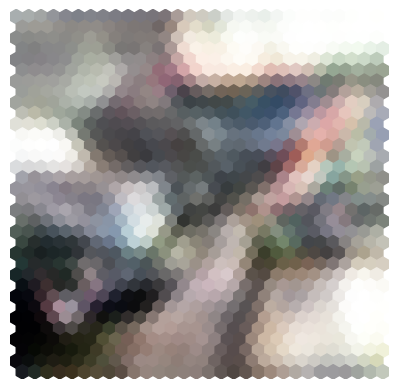}
    \includegraphics[width=\width]{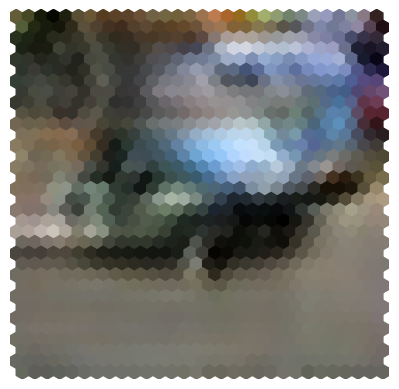}
    \includegraphics[width=\width]{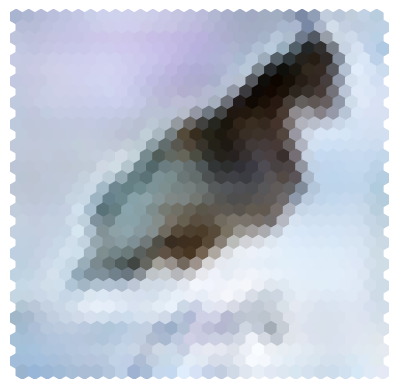}
    \includegraphics[width=\width]{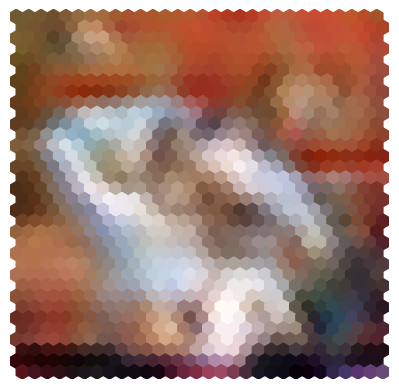}
    \includegraphics[width=\width]{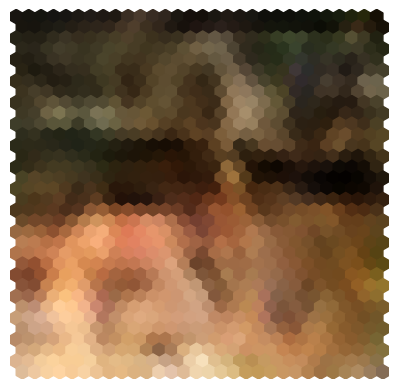}
    \vspace{0.1cm}

    \includegraphics[width=\width]{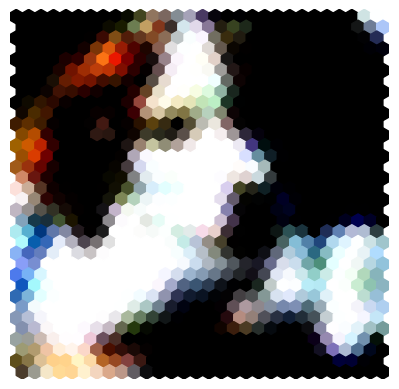}
    \includegraphics[width=\width]{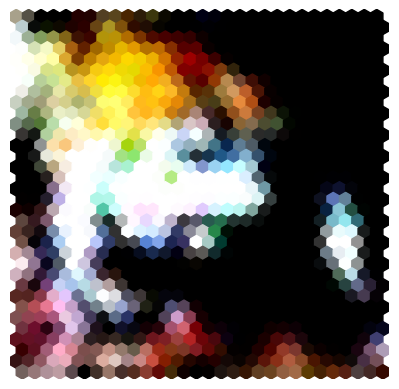}
    \includegraphics[width=\width]{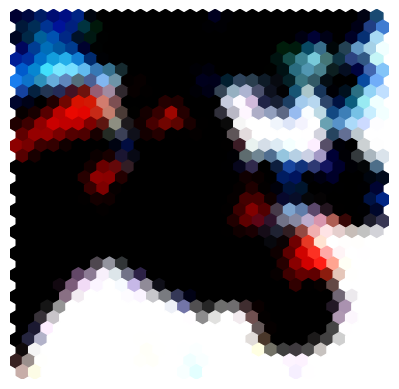}
    \includegraphics[width=\width]{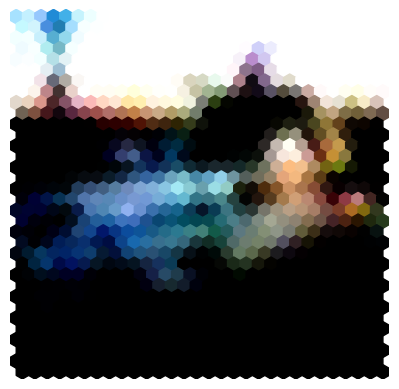}
    \includegraphics[width=\width]{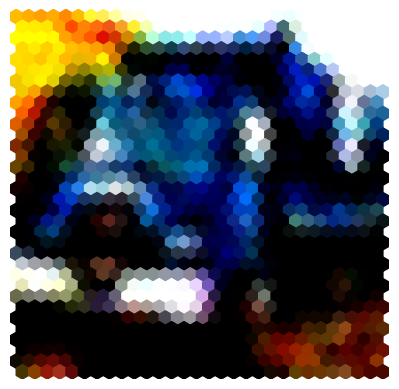}
    \hspace{0.2cm}
    \includegraphics[width=\width]{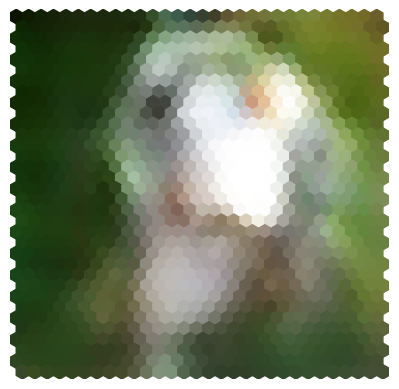}
    \includegraphics[width=\width]{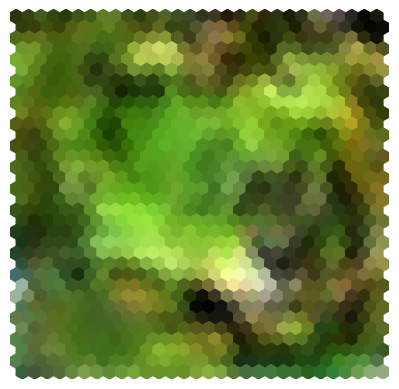}
    \includegraphics[width=\width]{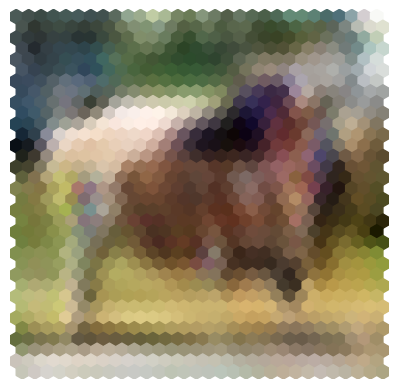}
    \includegraphics[width=\width]{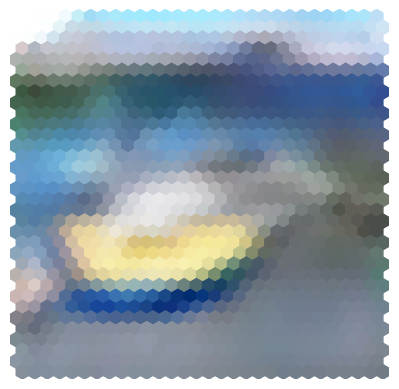}
    \includegraphics[width=\width]{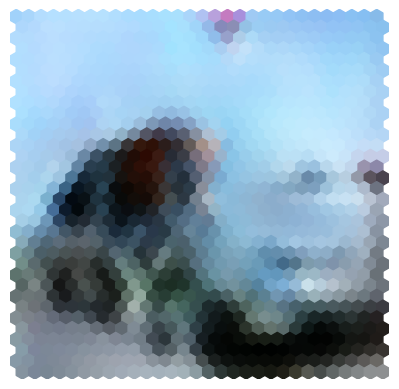}
    \vspace{-0.2cm}

    \rule[0.1cm]{\linewidth}{0.5pt}

    % COIL-100 hexagonal

    \includegraphics[width=\width]{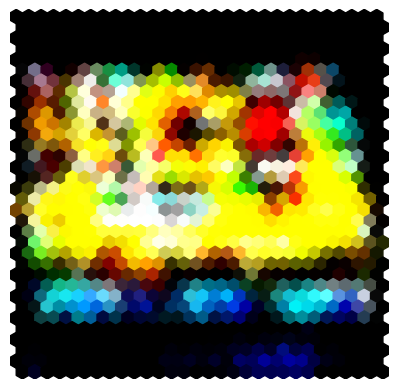}
    \includegraphics[width=\width]{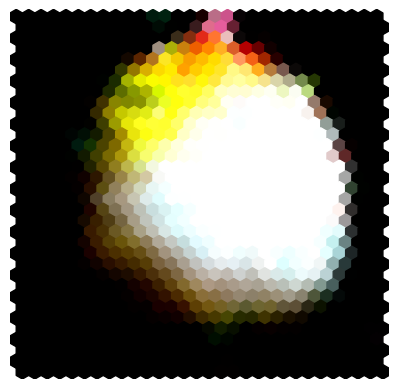}
    \includegraphics[width=\width]{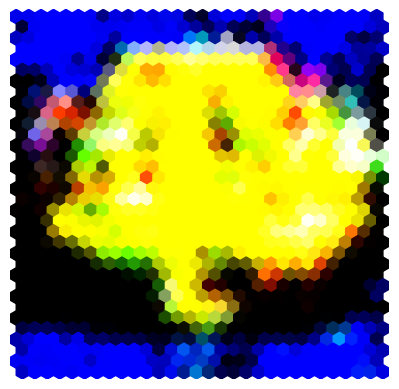}
    \includegraphics[width=\width]{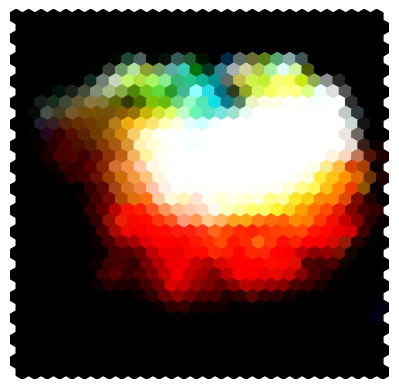}
    \includegraphics[width=\width]{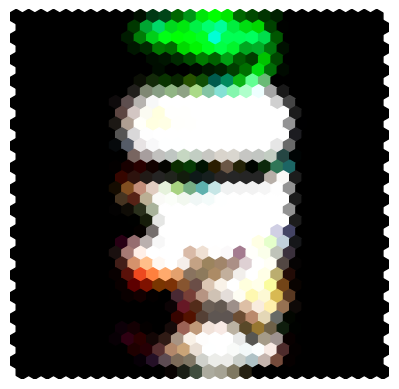}
    \hspace{0.2cm}
    \includegraphics[width=\width]{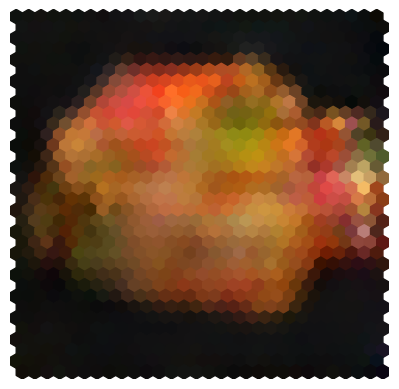}
    \includegraphics[width=\width]{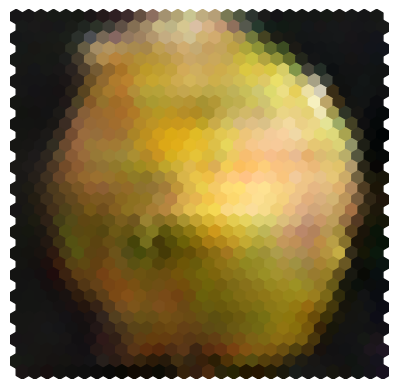}
    \includegraphics[width=\width]{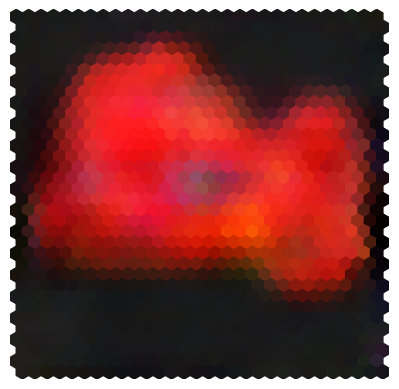}
    \includegraphics[width=\width]{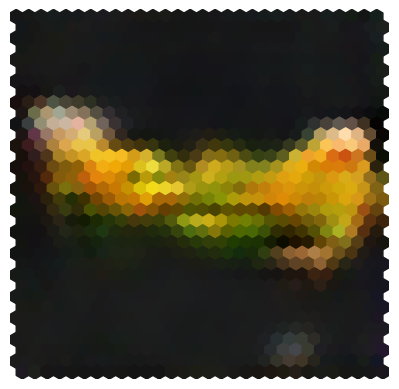}
    \includegraphics[width=\width]{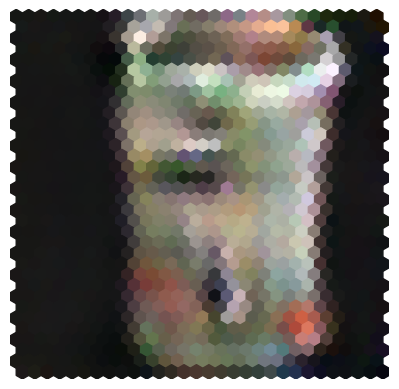}
    \vspace{0.1cm}

    \includegraphics[width=\width]{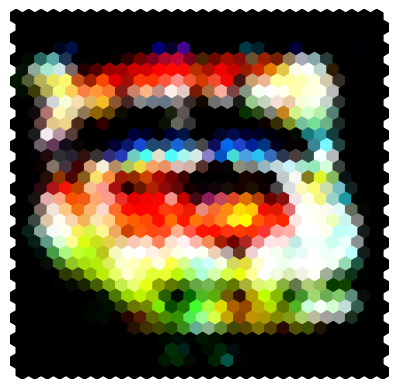}
    \includegraphics[width=\width]{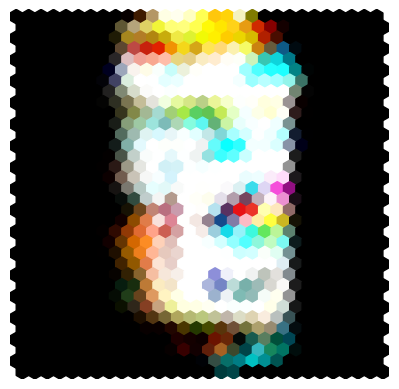}
    \includegraphics[width=\width]{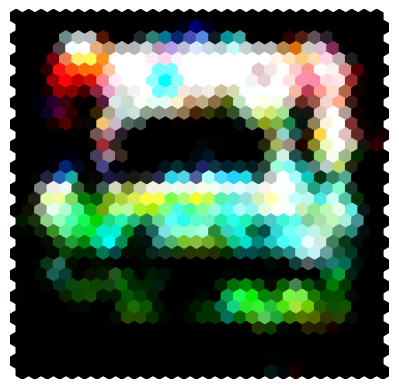}
    \includegraphics[width=\width]{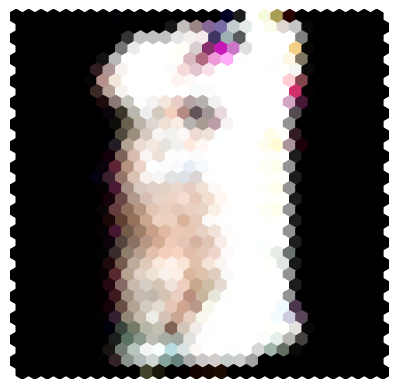}
    \includegraphics[width=\width]{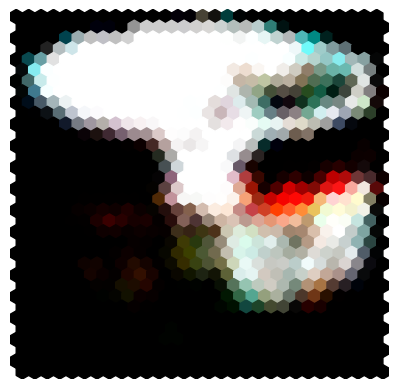}
    \hspace{0.2cm}
    \includegraphics[width=\width]{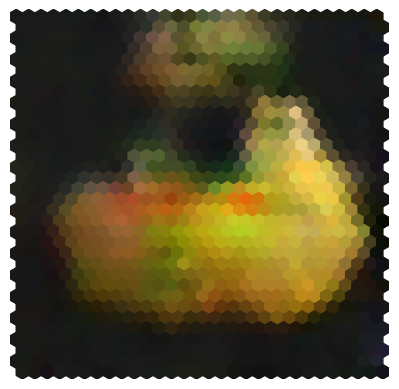}
    \includegraphics[width=\width]{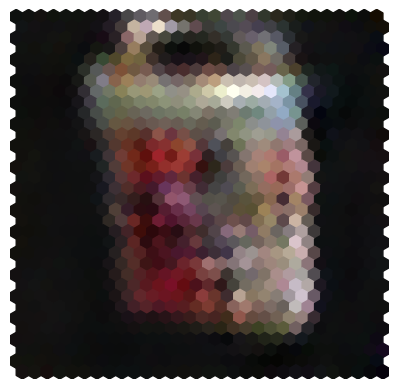}
    \includegraphics[width=\width]{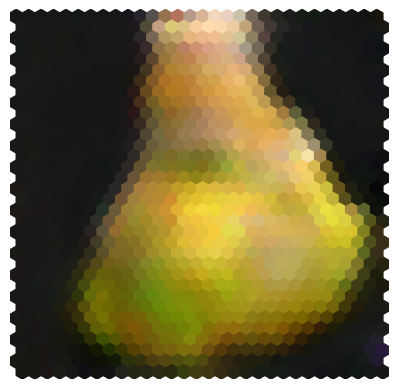}
    \includegraphics[width=\width]{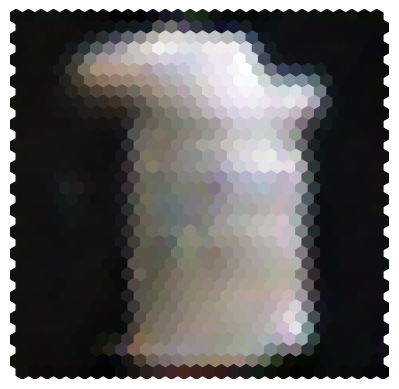}
    \includegraphics[width=\width]{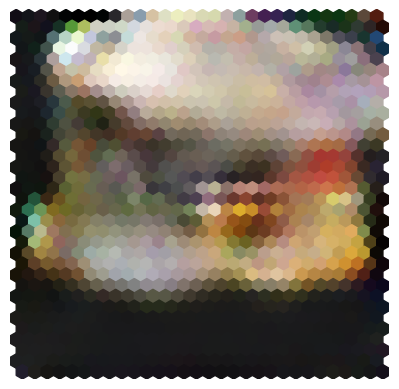}
    \vspace{0cm}

    \caption{Exemplary test results for S- and H-SWWAE (left column) as well as S- and H-ACGAN (right column) with the MNIST (top), CIFAR-10 (middle), and COIL-100 (bottom) data sets after 100 epochs of training.}
    \label{figure:test_results}
\end{figure}

\begin{table}[tb]
    \renewcommand{\arraystretch}{1.1}
    \centering

    \scalebox{0.9}{\begin{tabular}{|M{1.5cm}|M{1.75cm}|M{2cm}|M{2cm}|}
        \hline
        Data set & Test run & SWWAE & ACGAN \\
        \hline
        \noalign{\vskip 2pt}

        \hline
        \multirow{2}{*}{CIFAR-10} & square             & $37.3 \pm 2.4$          & $23.7 \pm 4.5$ \\
        \cline{2-4}
                                  & \textbf{hexagonal} & $\mathbf{39.1 \pm 2.5}$ & $\mathbf{24.8 \pm 4.1}$ \\
        \hline
        \multirow{2}{*}{COIL-100} & square             & $39.4 \pm 2.7$          & $25.3 \pm 3.9$ \\
        \cline{2-4}
                                  & \textbf{hexagonal} & $\mathbf{39.7 \pm 2.6}$ & $\mathbf{26.3 \pm 3.9}$ \\
        \hline
        \multirow{2}{*}{MNIST}    & square             & $44.2 \pm 1.9$          & $32.3 \pm 3.7$ \\
        \cline{2-4}
                                  & \textbf{hexagonal} & $\mathbf{45.1 \pm 1.7}$ & $\mathbf{33.6 \pm 4.2}$ \\
        \hline
    \end{tabular}}

    \caption{PSNR-based transformation efficiency test results after 100 epochs of training for one batch of images.}
    \label{table:test_results}
\end{table}

\section{Results, evaluation, and discussion}

%%%%%%%%%%%%%%%%%%%%%%%%%%%%%%%%%%%%% Text %%%%%%%%%%%%%%%%%%%%%%%%%%%%%%%%%%%%%
Current research that tries to combine principles from the domain of hexagonal image processing with machine learning often only provides application-specific test results with data which were either synthesized or captured using a hexagonal sensor. However, this contribution aims to provide a more general evaluation by assessing the quality of different S- and H-DNN models with square and hexagonal images.
%%%%%%%%%%%%%%%%%%%%%%%%%%%%%%%%%%%%% Text %%%%%%%%%%%%%%%%%%%%%%%%%%%%%%%%%%%%%

\subsection{Square and hexagonal deep neural networks}

%%%%%%%%%%%%%%%%%%%%%%%%%%%%%%%%%%%%% Text %%%%%%%%%%%%%%%%%%%%%%%%%%%%%%%%%%%%%
To enable a comparison of different S- and H-DNN models, we deployed two different kernel sizes for all convolutional layers and tested them against various data sets with an initial image resolution of $32 \times 32$ square and $34 \times 30$ hexagonal pixels. The realized architecture configurations for 10 classes are shown in Tables~\ref{table:SWWAE_model_configuration} and \ref{table:ACGAN_model_configuration}, resulting in $251\,931$ and $195\,946$ (S- and H-SWWAE) as well as $1\,149\,603$ and $1\,123\,706$ (S- and H-ACGAN) trainable parameters respectively. In comparison to the layer configurations as given by the original authors for both S-DNNs, the deployed kernel sizes were therefore substituted as shown. Our proposed training setup includes: the Glorot initializer for weight initialization, the Adam optimizer with a learning rate of $0.0002$ and exponential decay rates of $0.5$ and $0.999$, as well as a batch size of $100$. The SWWAE models were then trained by deploying the MSE-based transformation efficiency in~\eqref{equation:MSE} as loss function.
%%%%%%%%%%%%%%%%%%%%%%%%%%%%%%%%%%%%% Text %%%%%%%%%%%%%%%%%%%%%%%%%%%%%%%%%%%%%

\subsection{Quality assessment}

%%%%%%%%%%%%%%%%%%%%%%%%%%%%%%%%%%%%% Text %%%%%%%%%%%%%%%%%%%%%%%%%%%%%%%%%%%%%
To assess the quality of the generated images, we evaluated the CIFAR-10, COIL-100, and MNIST data sets using their given data set split ratios. Our S- and H-SWWAE as well as S- and H-ACGAN test results after 100 epochs of training are shown in Fig.~\ref{figure:test_results}. For the differences in image quality of square and hexagonally generated images, the transformation efficiencies for the respective data sets are shown in Table~\ref{table:test_results} using the peak signal-to-noise ratio (PSNR). These results were obtained by comparing the original square lattice format based images with the generated hexagonal ones. Whereas a comparison of all test results for the given models and their proposed hexagonal equivalents reveals increased efficiencies for both hexagonal models, they also result in a reduction of trainable parameters. The hexagonal models enable therefore the reduction of the models' complexity, while hexagonal layers can furthermore improve image quality.
%%%%%%%%%%%%%%%%%%%%%%%%%%%%%%%%%%%%% Text %%%%%%%%%%%%%%%%%%%%%%%%%%%%%%%%%%%%%

\section{Conclusion and outlook}

%%%%%%%%%%%%%%%%%%%%%%%%%%%%%%%%%%%%% Text %%%%%%%%%%%%%%%%%%%%%%%%%%%%%%%%%%%%%
Our test results highlight that H-DNNs for hexagonal image generation can outperform conventional DNNs, whereas an increase in test rates and the reduction of trainable parameters demonstrates their benefits. Future deep learning models can form the hexagonal equivalents to conventional DNNs, as further novel and application-specific models and data sets have to be developed and investigated.
%%%%%%%%%%%%%%%%%%%%%%%%%%%%%%%%%%%%% Text %%%%%%%%%%%%%%%%%%%%%%%%%%%%%%%%%%%%%

\section{Acknowledgment}

%%%%%%%%%%%%%%%%%%%%%%%%%%%%%%%%%%%%% Text %%%%%%%%%%%%%%%%%%%%%%%%%%%%%%%%%%%%%
The European Union and the European Social Fund for Germany partially funded this research.
%%%%%%%%%%%%%%%%%%%%%%%%%%%%%%%%%%%%% Text %%%%%%%%%%%%%%%%%%%%%%%%%%%%%%%%%%%%%

\bibliographystyle{IEEEbib}
\bibliography{library_without_pages}

\begin{thebibliography}{10}

\bibitem{Staunton1990}
Richard~C. Staunton and Neil Storey,
\newblock ``{A Comparison Between Square and Hexagonal Sampling Methods for
  Pipeline Image Processing},''
\newblock in {\em Optics, Illumination, and Image Sensing for Machine Vision
  IV}, 1990, vol. 1194.

\bibitem{Krizhevsky2012}
Alex Krizhevsky, Ilya Sutskever, and Geoffrey~E. Hinton,
\newblock ``{ImageNet Classification with Deep Convolutional Neural
  Networks},''
\newblock in {\em Advances in Neural Information Processing Systems 25}, 2012.

\bibitem{Szegedy2015}
Christian Szegedy, Wei Liu, Yangqing Jia, Pierre Sermanet, Scott Reed, Dragomir
  Anguelov, Dumitru Erhan, Vincent Vanhoucke, and Andrew Rabinovich,
\newblock ``{Going Deeper with Convolutions},''
\newblock in {\em The IEEE Conference on Computer Vision and Pattern
  Recognition (CVPR)}, 2015.

\bibitem{Bronstein2017}
Michael~M. Bronstein, Joan Bruna, Yann LeCun, Arthur Szlam, and Pierre
  Vandergheynst,
\newblock ``{Geometric Deep Learning: Going beyond Euclidean data},''
\newblock {\em IEEE Signal Processing Magazine}, vol. 34, no. 4, 2017.

\bibitem{Middleton2005}
Lee Middleton and Jayanthi Sivaswamy,
\newblock {\em {Hexagonal Image Processing: A Practical Approach}},
\newblock Springer, 2005.

\bibitem{Curcio1987}
C.~A. Curcio, K.~R.~Jr Sloan, O.~Packer, A.~E. Hendrickson, and R.~E. Kalina,
\newblock ``{Distribution of Cones in Human and Monkey Retina: Individual
  Variability and Radial Asymmetry},''
\newblock {\em Science}, vol. 236, no. 4801, 1987.

\bibitem{Hubel1968}
D.~H. Hubel and T.~N. Wiesel,
\newblock ``{Receptive fields and functional architecture of monkey striate
  cortex},''
\newblock {\em The Journal of Physiology}, vol. 195, no. 1, 1968.

\bibitem{Petersen1962}
Daniel~P. Petersen and David Middleton,
\newblock ``{Sampling and Reconstruction of Wave-Number-Limited Functions in
  N-Dimensional Euclidean Spaces*},''
\newblock {\em Information and Control}, vol. 5, no. 4, 1962.

\bibitem{Golay1969}
M.~J.~E. Golay,
\newblock ``{Hexagonal Parallel Pattern Transformations},''
\newblock {\em IEEE Transactions on Computers}, vol. C-18, no. 8, 1969.

\bibitem{Birch2007}
Colin P.~D. Birch, Sander~P. Oom, and Jonathan~A. Beecham,
\newblock ``{Rectangular and hexagonal grids used for observation, experiment
  and simulation in ecology},''
\newblock {\em Ecological Modelling}, vol. 206, no. 3-4, 2007.

\bibitem{Sahr2003}
Kevin Sahr, Denis White, and A.~Jon Kimerling,
\newblock ``{Geodesic Discrete Global Grid Systems},''
\newblock {\em Cartography and Geographic Information Science}, vol. 30, no. 2,
  2003.

\bibitem{Ambrosio2001}
M.~Ambrosio, C.~Aramo, F.~Bracci, P.~Facal, R.~Fonte, G.~Gallo, E.~Kemp,
  G.~Matthiae, D.~Nicotra, P.~Privitera, G.~Raia, E.~Tusi, and G.~Vitali,
\newblock ``{The Camera of the Auger Fluorescence Detector},''
\newblock {\em IEEE Transactions on Nuclear Science}, vol. 48, no. 3, 2001.

\bibitem{Neeser2000}
W.~Neeser, M.~Bocker, P.~Buchholz, P.~Fischer, P.~Holl, J.~Kemmer, P.~Klein,
  H.~Koch, M.~Locker, G.~Lutz, H.~Matthay, L.~Struders, M.~Trimpl, J.~Ulrici,
  and N.~Wermes,
\newblock ``{The DEPFET pixel BIOSCOPE},''
\newblock {\em IEEE Transactions on Nuclear Science}, vol. 47, no. 3, 2000.

\bibitem{Theussl2001}
T.~Theussl, T.~Moller, and M.~E. Groller,
\newblock ``{Optimal Regular Volume Sampling},''
\newblock in {\em Proceedings Visualization, 2001. VIS '01.}, 2001.

\bibitem{Ke2018}
Jintao Ke, Hai Yang, Hongyu Zheng, Xiqun Chen, Yitian Jia, Pinghua Gong, and
  Jieping Ye,
\newblock ``{Hexagon-Based Convolutional Neural Network for Supply-Demand
  Forecasting of Ride-Sourcing Services},''
\newblock {\em IEEE Transactions on Intelligent Transportation Systems}, 2018.

\bibitem{Shilon2019}
I.~Shilon, M.~Kraus, M.~B{\"{u}}chele, K.~Egberts, T.~Fischer, T.~L. Holch,
  T.~Lohse, U.~Schwanke, C.~Steppa, and S.~Funka,
\newblock ``{Application of Deep Learning methods to analysis of Imaging
  Atmospheric Cherenkov Telescopes data},''
\newblock {\em Astroparticle Physics}, vol. 105, 2019.

\bibitem{Steppa2019}
Constantin Steppa and Tim~L. Holch,
\newblock ``{HexagDLy—Processing hexagonally sampled data with CNNs in
  PyTorch},''
\newblock {\em SoftwareX}, vol. 9, 2019.

\bibitem{Schlosser2019_ICMLA}
Tobias Schlosser, Michael Friedrich, and Danny Kowerko,
\newblock ``{Hexagonal Image Processing in the Context of Machine Learning:
  Conception of a Biologically Inspired Hexagonal Deep Learning Framework},''
\newblock in {\em 2019 18th IEEE International Conference on Machine Learning
  and Applications (ICMLA)}, 2019.

\bibitem{Burt1980}
Peter~J. Burt,
\newblock ``{Tree and Pyramid Structures for Coding Hexagonally Sampled Binary
  Images},''
\newblock {\em Computer Graphics and Image Processing}, vol. 14, no. 3, 1980.

\bibitem{Fitz1996}
A.~P. Fitz and R.~J. Green,
\newblock ``{Fingerprint classification using a hexagonal fast fourier
  transform},''
\newblock {\em Pattern Recognition}, vol. 29, no. 10, 1996.

\bibitem{Her1994}
I.~Her and C.~T. Yuan,
\newblock ``{Resampling on a Pseudohexagonal Grid},''
\newblock {\em CVGIP: Graphical Models and Image Processing}, vol. 56, no. 4,
  1994.

\bibitem{Zhao2015}
Junbo Zhao, Michael Mathieu, Ross Goroshin, and Yann LeCun,
\newblock ``{Stacked What-Where Auto-encoders},''
\newblock in {\em 4th International Conference on Learning Representations
  (ICLR)}, 2016.

\bibitem{Odena2017}
Augustus Odena, Christopher Olah, and Jonathon Shlens,
\newblock ``{Conditional Image Synthesis with Auxiliary Classifier GANs},''
\newblock in {\em 34th International Conference on Machine Learning (ICML)},
  2017.

\end{thebibliography}

\end{document}